\documentclass[iicol,referee,pdflatex,sn-apa]{sn-jnl}

\usepackage{graphicx}%
\usepackage{multirow}%
\usepackage{amsmath,amssymb,amsfonts}%
\usepackage{amsthm}%
\usepackage{mathrsfs}%
\usepackage[title]{appendix}%
\usepackage{xcolor}%
\usepackage{textcomp}%
\usepackage{manyfoot}%
\usepackage{booktabs}%
\usepackage{algorithm}%
\usepackage{algorithmicx}%
\usepackage{algpseudocode}%
\usepackage{listings}%

\usepackage{subcaption}
\usepackage{mathtools}
\usepackage{url}
\usepackage{tabularray}
\usepackage{orcidlink}

\theoremstyle{thmstyleone}%
\theoremstyle{thmstyletwo}%

\theoremstyle{thmstylethree}%

\begin{document}

\title[Solving the Aircraft Disassembly Scheduling Problem]{Solving the Aircraft Disassembly Scheduling Problem}


\author*[1]{\fnm{Charles} \sur{Thomas} \orcidlink{0000-0002-7360-5372}}\email{charles.thomas@uclouvain.be}

\author*[1]{\fnm{Pierre} \sur{Schaus} \orcidlink{0000-0002-3153-8941}}\email{pierre.schaus@uclouvain.be}

\affil*[1]{\orgdiv{Institute of Information and Communication Technologies, Electronics and Applied Mathematics (ICTEAM)}, \orgname{UCLouvain}, \orgaddress{\street{Place Sainte Barbe 2/L5.02.01}, \city{Louvain-la-Neuve}, \postcode{1348}, \country{Belgium}}}


\abstract{
Dismantling aircrafts reaching their end of life is a complex endeavour that is necessary in terms of sustainability but yields small income margins for air transport companies.
An efficient scheduling of the disassembly procedure is thus crucial to ensure the profitability of the process and incentivize practice.
This is a large scheduling problem that involves thousands of tasks and many different constraints: 
Extracting parts that are destined to be reused requires technicians with specific certifications and equipment.
Extraction operations might be subject to precedence relations.
Furthermore, the aircraft must be kept balanced during the whole process.
Finally, some of the locations of the aircraft have a limited space that caps the number of technicians able to work there concurrently.
This article presents the problem in details and proposes two approaches to solve the problem: a Constraint Programming model and a MIP model.
The models are tested on instances of varying sizes involving up to 1450 tasks, which are based on real operational data provided by an industrial partner.
}




\maketitle

\section*{Acknowledgments}

Charles Thomas contribution was funded by the Walloon Region (Belgium) as part of the Planum project.
We thank Sabena Engineering for allowing the diffusion of an anonymized version of the dataset provided.

\section{Introduction}
\label{sec:intro}

The air transport industry is a growing sector that has rebounded since the COVID-19 pandemic.
Despite a recent slowing down, the growth of the sector is projected to continue in the future with an annual production of around 2000 new aircrafts per year (\cite{iata2025global}).
While the global number of planes in activity is increasing, so too is the number of aircrafts that reach their end of life (EOL) phase.
Dealing with these aircrafts in an economically beneficial way while limiting their environmental impact is an important challenge for the air transport industry.

Recent studies (see \cite{asmatulu2013recycling,sabaghi2016towards,scheelhaase2022economic,habib2025current}) indicate that, while aircraft recycling provides environmental and economic advantages, the resulting profit margins are strongly influenced by the condition of the recycled aircraft.
The value of an EOL aircraft usually ranges between USD 1 million and USD 3 million.
However, this value is conditional to the state of the aircraft and whether it is in order of maintenance.
Currently, most of the value come from the reselling of parts extracted during the disassembly process, of which engine components are the most valuable and may account up to 80\% of the value recovered.

While valuable in theory, materials are more difficult to reuse due to the mixing of several components into alloys and the increased use of polymer materials.
Current recycling processes for both metals and polymers require a substantial amount of energy and must deal with the presence of impurities such as undesirable elements from paints and isolation, which make their recycling costly.

Nevertheless, improvements in the recycling processes as well as environmental and regulatory considerations are driving airlines towards investing in the recycling of EOL aircrafts.
Current market analyses (\cite{kpmg2024circularity,gmi2024aircraft,fortune2025commercial}) estimate the global aircraft recycling market at USD 5 to 7 billion with 400 to 450 aircrafts dismantled annually.
This market is projected to grow to around USD 14 billion by 2034 with more than 1000 aircraft expected to be retired annually.

In this context, the Planum research project aims at establishing an industry around the dismantling and treatment of end of life aircrafts in Wallonia (Belgium).
As part of this project, efforts have been made to improve the efficiency of the dismantling process.
One of these research areas concerns the scheduling of the disassembly tasks.

The dismantling of an aircraft consists of the following steps:
First, parts and elements that can be reused or individually recycled are removed from the aircraft.
Additionally, several pollutants also need to be removed in this phase.
Once this is done, the carcass of the aircraft remains.
It is then cut and shredded into small material pieces that are then sorted and treated to be either recycled or disposed of.

This paper studies the scheduling of the disassembly tasks that occur in the first half of this process, up to the sectional cutting and shredding of the aircraft.
Disassembly tasks mostly consist in the removal of pieces and materials from the aircraft but also include checks, activation and deactivation of systems and preparation work such as opening access panels and setting up scaffoldings.
Tasks have variable durations going from 15 minutes to 16 hours. The order in which tasks are performed may be restricted by precedence constraints.

Tasks each require a given number of technicians that must be assigned in addition to the scheduling.
Some tasks have certification requirements, which limit the technicians that may be assigned.
Some technicians may also be unavailable at some times during the scheduling process.
Additionally, place is limited in some parts of the aircraft which constrains the number of technicians able to work simultaneously in these locations and thus tasks that can be performed concurrently.
Finally, the balance of the aircraft must be maintained during the whole disassembly process, which puts additional constraints on the order in which tasks where a certain quantity of mass is removed are done.
The objective of the problem is to minimize the \textit{makespan}, i.e. the time at which the last task of the schedule ends.

The problem studied in this paper is based on a real use case provided by an industrial partner.
The data used for the experiments is based on the disassembly of a Boeing B737 plane, which consists in around 1500 tasks to schedule.

This article is an extension of the work presented in \cite{thomas2024constraint}.
The aircraft disassembly scheduling problem is presented with additional details, including figures and an example.
In addition to the Constraint Programming (CP) approach presented in \cite{thomas2024constraint}, a Mixed Integer Programming (MIP) approach is proposed.
Both approaches are then compared on several instances generated based on real data from an industrial partner.
The data is also presented in details and its characteristics analyzed.
Several variations of the problem are considered where some constraints are deactivated selectively to examine their impact.

Section \ref{sec:rw} discusses related work on similar scheduling problems arising in a dismantling context.
Section \ref{sec:prob} presents the problem and gives a small example.
The CP model is presented and explained in Section \ref{sec:cp}.
The MIP model is shown in Section \ref{sec:mip}.
Section \ref{sec:xp} presents the experiments done and their results.
Finally, Section \ref{sec:ccl} offers some closing remarks as well as possible avenues for future research.

\section{Related work}
\label{sec:rw}

The aircraft disassembly scheduling problem is a variation of the Resource Constrained Project Scheduling Problem (RCPSP) \citep{Ozdamar1995,Brucker1999}, which consists in scheduling a series of tasks consuming several resources under precedence constraints.
The objective is to find a feasible schedule that minimizes the makespan of the tasks.
This problem is NP-hard \citep{garey1975complexity}.
Several variants of the problem exist \citep{Hartmann2022}.
The closest one to our problem is probably the Multi-Skill Project Scheduling Problem (MSPSP) introduced in \cite{bellenguez2004lower}.
It consists in scheduling tasks and assigning workers with different skill levels to them.
It is essentially a relaxed version of the Aircraft Disassembly Scheduling problem discussed in this article without capacity or balance constraints.
In \cite{young2017constraint}, the authors use a CP model to solve several instances of the MSPSP with up to 60 tasks, 19 workers and 15 different skills.
In \cite{polo2023heuristic}, a combination of heuristic and metaheuristic approaches is used to solve a variant of the MSPSP with 50 tasks.

Regarding the balancing constraints, the chemical tanker scheduling and tank allocation problem is also concerned with maintaining ship stability and trim equilibrium (see \cite{vilhelmsen2016heuristic}).
\cite{le2025aircraft} introduced resource-constrained assembly line balancing for aircraft manufacturing  where resource utilization should remain relatively balanced across production periods to ensure a stable workflow.
Cargo optimization are also problems embedding center of gravity and load distribution constraints as core requirements (see \cite{limbourg2012automatic}).
More generally the balance constraints in RCPSP problems can be modeled using cumulative or storage resources (see \cite{neumann2003project}).

Dismantling an aircraft is the opposite of assembling it.
\cite{pucel2024constraint} developed a constraint programming model for assembly line balancing of aircrafts with workstations, work zones within workstations, and cumulative resources shared across workstations.
This model represents an extension of RCPSP with spatial divisions (workstations and zones) where capacity constraints limit concurrent operations in each spatial unit.
Similarly, \cite{brimberg1996scheduling} presents an aircraft maintenance scheduling problem with up to 350 tasks where extra constraints are used to take into account the limited physical space in some parts of the aircraft.
More generally, the capacity of certain locations within the aircraft—expressed as the maximum number of technicians who can work simultaneously in a given area—can be interpreted as a standard renewable resource, and represents a special case of spatial resources (see \cite{de1998resource}).

Other publications are related to the problem studied in this paper:
In \cite{shan2017adaptive}, the authors propose a genetic algorithm to solve an aircraft assembly RCPSP.
The authors of \cite{borsato2022integer} propose an integer programming approach to schedule aircraft engine assembly lines that also involves workers with several skills on up to 100 tasks.
In \cite{niu2023short} the authors propose an approach to schedule technicians on short-term aviation maintenance processes (up to 48h).
In \cite{srinivasan1999selective,zhong2011disassembly,camelot2013decision,dayi2016lean} different approaches are studied to solve problems linked to aircraft disassembly by finding optimal sequences to access specific components based on spatial and geometrical data.

Several CP approaches have also been proposed for problems linked to disassembly scheduling:
In \cite{lee2002disassembly}, a disassembly problem with capacity constraints is studied.
The stochastic aspects of disassembly processes are studied in \cite{bentaha2013chance} and \cite{tian2013chance}.
In \cite{zwingmann2008optimal,edis2021constraint,kizilay2022novel,hubner2021solving} several MILP and CP models are proposed to solve disassembly problems but are only able to solve instances up to 150-300 tasks.

In contrast, the instances tackled in this paper go up to around 1500 tasks which is substantially larger than most of the related work on the RCPSP.
Furthermore, despite several papers presenting similar constraints to the ones present in this work, none of them considers the combination of skills, balance and capacity constraints at the same time.


\section{Problem}
\label{sec:prob}

This section presents the aircraft disassembly problem, introduces a formal definition of the problem (Subsection \ref{sec:def}) and illustrates the problem with an example (Subsection \ref{sec:ex1}).

The Aircraft Dismantling Problem (ADP) consists in scheduling the set of tasks performed during the disassembly of the aircraft.
Technicians must be assigned to tasks following strict requirements:
Each task involves a specific number of technicians.
For tasks relative to the retrieval of parts that are to be reused, at least one technician in the team must have a specific certification.
Some technicians may have periods of unavailability during the project span.

In addition, space is limited in some parts of the plane, which limits the number of technicians able to work concurrently in these places.
For example, a cargo hold may accommodate at most two or three technicians at the same time, which limits the number of tasks that can be done in parallel there.
Thus, the plane is divided into locations, which each have an occupancy limit corresponding to the maximum number of technicians allowed to work there at the same time.

There are precedences between some operations as retrieving some parts my require the prior removal of other parts or the setup of dedicated equipment.
Additionally, some groups of tasks must be performed at specific points in the dismantling process.
For example, when the aircraft is received, a series of tasks consisting of several integrity and performance tests are done prior to any dismantling operation.
These constraints are also introduced as precedences.

Finally, the plane must be kept balanced during the whole disassembly process by ensuring that the difference of mass between its extremities does not overstep given thresholds.
To do so, two balance axis are considered:
The first axis opposes the front of the aircraft with its rear.
The second balance axis opposes both wings of the aircraft.
At any time of the planning, the difference of mass between the front and the rear of the aircraft as well as the difference between the left and right wings must not exceed given thresholds.

The objective is to minimize the total time taken by the whole extraction process.
This is represented by a \textit{makespan} value that corresponds to the time step at which the last operation finishes.

\subsection{Formal definition}
\label{sec:def}
In formal terms, the problem is defined as such:
The set of all tasks to perform is denoted $\mathcal{T}$.
Each task $i \in \mathcal{T}$ is defined by the following elements:
a duration needed to perform the task $d_i$,
a location $l_i$ where the task takes place,
an occupancy $\tau_i$, which corresponds to the number of technicians mobilized by the task,
a mass removed $m_i$,
a set of precedences $\mathcal{P}_i$ referring to other tasks that must end before the start of the task and
a set of requirements needed to perform the task $\mathcal{Q}_i$.
Each requirement $q \in \mathcal{Q}_i$ of this set is a pair $(c_{iq}, n_{iq})$ where
$c_{iq} \in \mathcal{C}$ is the skill (or certification) required and
$n_{iq}$ indicates the number of technicians with this skill needed.

Note that some tasks may not have a mass value or specific requirements, in which case, these fields are empty.
Furthermore, some tasks do not have a specific location or span several locations.
Thus their $l_i$ attribute may be empty.
Such tasks either do not count towards the occupancy constraint or mobilize the whole plane or technical team.
In the latter case, these tasks must occur at a specific phase in the dismantling process, which is enforced by precedence constraints.

For example, after the reception of the aircraft, there is a series of basic maintenance checks, followed by a test run of the engines \cite{skybrary2025Aircraft}.
This procedure consist in a chain of tasks: fueling the plane, towing it to the test area, performing the power run, towing it back and purging its tanks.
These tasks mobilize all the plane and must be performed in this strict order as enforced by precedences.
Afterwards, the actual disassembly can start.
Additional precedence constraints are used to make sure that any task that is part of the basic maintenance check must be done before the fueling of the plane, which marks the start of the test run procedure.
Similarly, any disassembly task must be done after the purging of the tank at the end of the procedure.

All available technicians are part of the set $\mathcal{R}$.
Each technician $j \in \mathcal{R}$ is associated with
a set of skills $\mathcal{C}_j$
and a set of unavailabilities $u \in \mathcal{U}_j$ consisting of time windows $[s_{ju},e_{ju}]$ when the technician is not available.

A set of locations $\mathcal{L}$ contains all the locations where operations can take place.
Each location $l \in \mathcal{L}$ is associated to a capacity $k_l$ that indicates the maximum number of technicians that can work simultaneously in this location and
optionally a zone $z_l$ that corresponds to one of the balance zones of the aircraft.
There are four balance zones in total: \texttt{Aft} and \texttt{Fwd} that correspond to the rear and front of the aircraft and \texttt{Left} and \texttt{Right} that correspond to the wings.

Two global parameters, $B^{af}$ and $B^{lr}$, indicate the maximum difference of mass allowed at any point in the planning between the \texttt{Aft} and \texttt{Fwd} zones and the \texttt{Left} and \texttt{Right} zones respectively.

The objective is to minimize the makespan under the following constraints:
\begin{enumerate}
    \item All the technicians needed for a task must be allocated during its whole duration.
    \item A technician cannot be allocated to different operations at the same time;
    \item A technician cannot be scheduled during its unavailabilities;
    \item Precedences between tasks must be respected;
    \item All the requirements needed for a task must be met.
    \item The capacity $k_l$ of a location must not be overloaded at any time;
    \item The difference of mass between the \texttt{Aft} and \texttt{Fwd} zones cannot overstep the balance parameter $B^{af}$ at any time during the planning;
    \item The difference of mass between the \texttt{Left} and \texttt{Right} zones cannot overstep the balance parameter $B^{lr}$ at any time during the planning;
\end{enumerate}

For balance constraints, we consider that the weight change induced by a task happens at the start time of the task.
Additionally, we denote $\Theta$ the global planning horizon which is equal to the sum of all tasks durations.
Finally, tasks are non-preemptive, meaning that, once started they cannot be interrupted and restarted at a later time. Furthermore, technicians are assigned to tasks for their whole duration and cannot switch tasks before the end of their current task.

Note that this last consideration, combined with the fact that technicians can have multiple skills as well as their own unavailability, makes infeasible any approach that exploits symmetries by grouping technicians into a single capacitated resource.
For example, let us consider two technicians: one available from time 0 to time 10 and another available from time 8 to time 15.
With an approach grouping these technicians into a single capacitated resource, the capacity would be higher or equal to 1 between time 0 and time 15. This would allow us to schedule a task of duration 15 requiring 1 technician in this time interval, which is not feasible in practice.

\subsection{Example}
\label{sec:ex1}

\begin{table*}
    \centering
    \begin{tabular}{|c|l|lrrlrl|}
    \hline
    Task  & Description                & $l_i$     & $d_i$ & $\tau_i$ & $\mathcal{Q}_i$ & $m_i$ & $\mathcal{P}_i$ \\ \hline
    A    & Empty Fuel Tanks           &           & 2     & 1        &                 &       &                 \\
    B    & Rem. Pilot Seat            & Cockpit   & 2     & 2        &                 &       & A               \\
    C    & Rem. Copilot Seat          & Cockpit   & 2     & 2        &                 &       & A               \\
    D    & Rem. Flight Controls Panel & Cockpit   & 3     & 1        & (B1,1)          &       & B,C             \\
    E    & Rem. L. Engine Thruster    & L. Wing   & 3     & 2        & (B2,1)          & 500   & A               \\
    F    & Rem. R. Engine Thruster    & R. Wing   & 3     & 2        & (B2,1)          & 500   & A               \\
    G    & Rem. L. Engine             & L. Wing   & 4     & 3        & (B2,1)          & 1200  & E               \\
    H    & Rem. R. Engine             & R. Wing   & 4     & 3        & (B2,1)          & 1200  & F               \\ \hline
    \end{tabular}
    \caption{Tasks of Example \ref{sec:ex1}}
    \label{tab:extsk}
\end{table*}

\begin{figure*}
    \centering
    \includegraphics{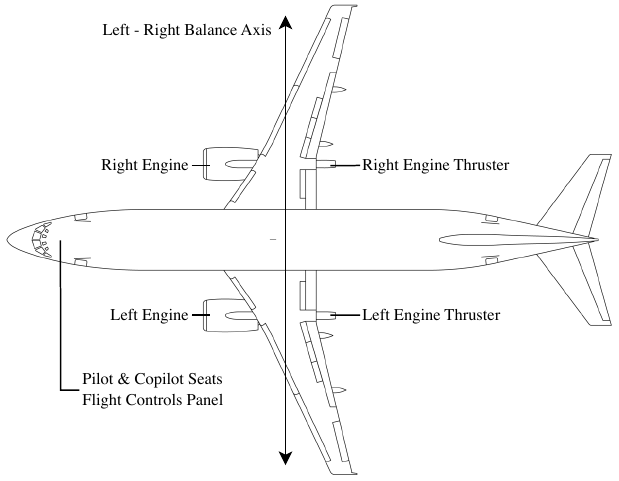}
    \caption{Location of the tasks of Example \ref{sec:ex1}}
    \label{fig:exloc}
\end{figure*}

Let us consider a small example with 8 dismantling tasks:
Task A consists in emptying the fuel tanks of the plane. It must be done at the start of the dismantling process, before the other tasks.
Tasks B and C consist in removing the Pilot and Copilot seats in the cockpit of the plane. Two technicians are required for each of these tasks.
Task D involves removing the flight controls panel in the cockpit. The pilot and copilot seats must be removed before in order for the technician to access this panel. This part can be reused and thus the task requires a technician with a B1 certification.
Tasks E and F consist in removing the thrusters on the left and right engine respectively. At least one technician with the B2 certification must be in the team for both of these tasks.
Finally, tasks G and H deal with the removal of the left and right engines. Each of them requires a team of 3 technicians, including 1 with a B2 certification. The thrusters must be removed prior to the removal of the engines.
The tasks and their characteristics are shown in Table \ref{tab:extsk}.

\begin{figure}
    \centering
    \includegraphics{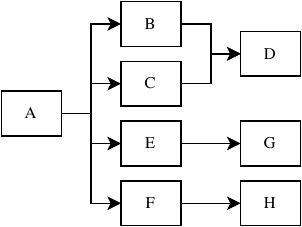}
    \caption{Precedences between the tasks of Example \ref{sec:ex1}}
    \label{fig:exprec}
\end{figure}

\begin{figure*}
    \centering
    \includegraphics[width=\textwidth]{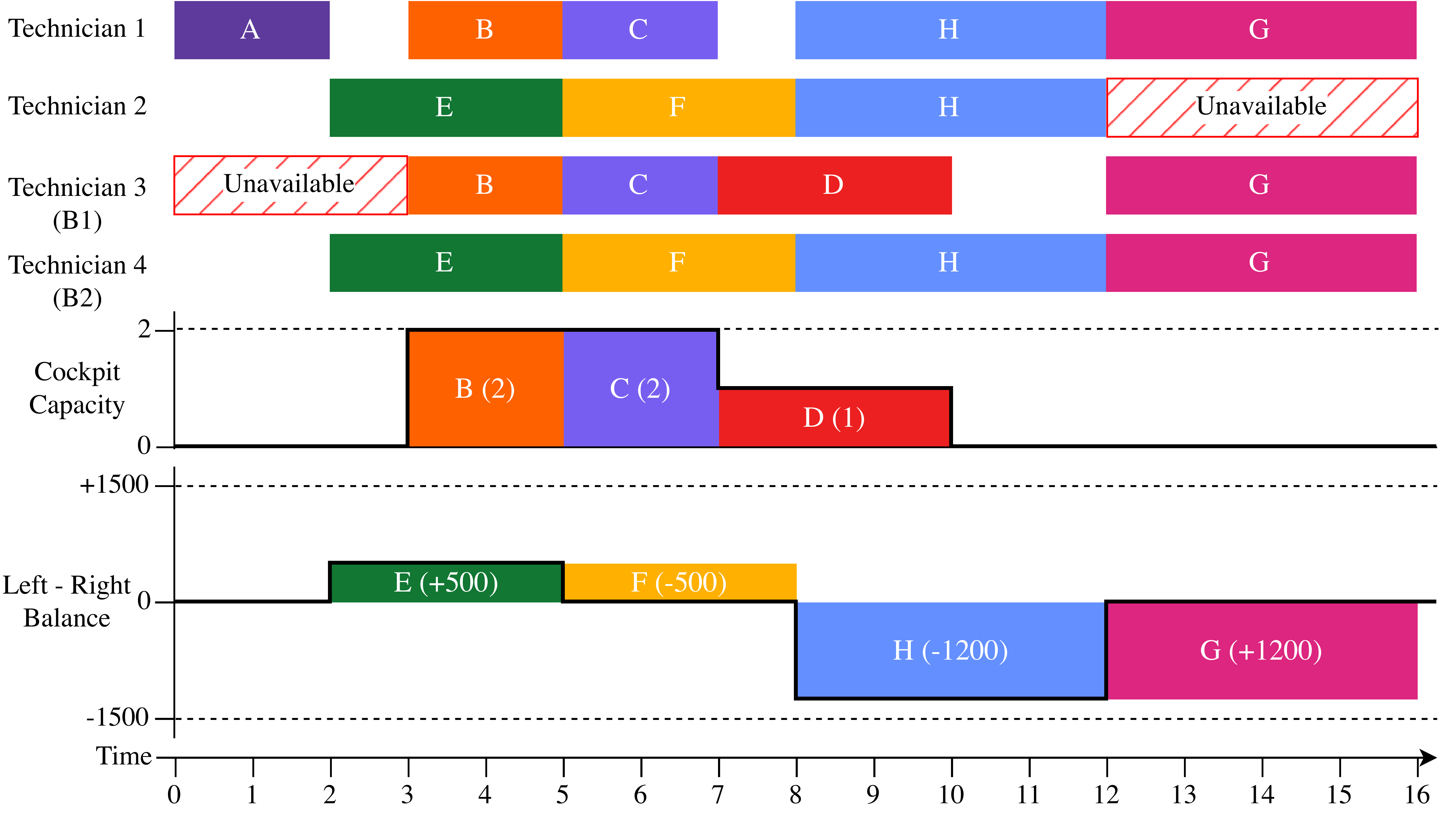}
    \caption{Illustration of a solution for Example \ref{sec:ex1}}
    \label{fig:exgant}
\end{figure*}

Only the cockpit location has a capacity constraint, allowing at most two technicians to be present simultaneously. The balance difference along the left-right axis must not exceed 1500.
Figure \ref{fig:exloc} shows the location of the tasks.
Note that Task A (Empty Fuel Tanks) has no location.
Figure \ref{fig:exprec} shows the precedences between tasks.

We have a team of 4 technicians.
Technician 2 is unavailable in the time interval $[12,\Theta]$.
Technician 3 has a B1 certification and is unavailable in the time interval $[0,3]$.
Technician 4 has a B2 certification.


A possible solution is shown in Table \ref{tab:exsol}.
Its makespan is 16.
This is an optimal solution.
\begin{table}[h]
    \centering
    \begin{tabular}{|c|rrl|}
    \hline
    Task & $s_i$ & $e_i$ & Technicians \\ \hline
    A    & 0     & 2     & 1           \\
    B    & 3     & 5     & 1,3         \\
    C    & 5     & 7     & 1,3         \\
    D    & 7     & 10    & 3           \\
    E    & 2     & 5     & 2,4         \\
    F    & 5     & 8     & 2,4         \\
    G    & 12    & 16    & 1,3,4       \\
    H    & 8     & 10    & 1,2,4       \\ \hline
    \end{tabular}
    \caption{Possible solution for Example \ref{sec:ex1}}
    \label{tab:exsol}
\end{table}
Figure \ref{fig:exgant} illustrates the solution.
The top part shows the assignations of the technicians.
The middle part shows the occupancy of the cockpit over time.
Tasks B, C and D cannot be done simultaneously due to the occupancy constraint.
The bottom part shows the evolution of the balance over the left right axis during the dismantling. Note that we must alternate between tasks on the left and right of the plane to satisfy this balance constraint.

\section{CP Model}
\label{sec:cp}

This section presents the constraint programming model proposed to solve the problem.
Subsection \ref{sec:cpb} provides a brief background on constraint programming and introduces several modeling concepts used in the model.
The variables used in the model are introduced in Subsection \ref{sec:cpv}.
Subsection \ref{sec:cpc} presents the model and details the constraints used.

\subsection{Background}
\label{sec:cpb}

Constraint Programming (CP) is a powerful paradigm for modeling and solving complex scheduling problems (see \cite{laborie2018ibm}).
Modern CP solvers provide specialized modeling constructs that enable the natural representation of scheduling problems through \textit{optional interval variables} and \textit{cumulative functions} introduced in \cite{laborie2008reasoning} and \cite{laborie2012interval}.

An \textit{interval variable} represents a time interval during which an activity is carried out.
Each interval variable $a$ is characterized by three attributes: a start time $s_a$, an end time $e_a$, and a duration $d_a$, where the relationship $s_a + d_a = e_a$ holds. 
An \textit{optional interval variables} enables the interval variable to be present or absent state. 
Several global constraints are defined on optional interval variables:
\begin{itemize}    
    \item The constraint $\textit{alternative}(a, \{a_1, \ldots, a_n\})$ ensures that if interval $a$ is present, then exactly one interval from $\{a_1, \ldots, a_n\}$ is present, and that $a$ starts and ends together with the chosen interval. 
    If $a$ is absent, all intervals in $\{a_1, \ldots, a_n\}$ are absent.
    
    \item The constraint $\textit{alternative}(a, \{a_1, \ldots, a_n\}, c)$ ensures that if $a$ is present, exactly $c$ intervals from $\{a_1, \ldots, a_n\}$ are present, with all selected intervals having the same start and end as $a$.
    
    \item The $\textit{noOverlap}(\{a_1, \ldots, a_n\})$ constraint enforces that no two present intervals in the set overlap in time.
\end{itemize}

\begin{figure}[h!]
    \centering
    \begin{tikzpicture}[>=latex, thick, scale=0.8, transform shape]

        \def\h{0.9}
        
        \draw[->] (0,0) -- (3.6,0);
        \draw[->] (0,0) -- (0,1.6);
        
        \draw (0,0) -- (0.8,0) -- (0.8,\h) -- (2.2,\h) -- (2.2,0) -- (3.4,0);
        
        \draw[fill=white] (0.8,-0.15) rectangle (2.2,-0.45);
        \node at (1.5,-0.3) {$a$};
        
        \node[right] at (3.5,\h) {$h$};
        \node[right] at (3.5,0) {$h$};
        \node at (1.5,1.3) {\textit{pulse}$(a,h)$};
        
        \draw[dotted] (0,\h) -- (3.4,\h);
        \draw[dotted] (0,0) -- (3.4,0);

        \begin{scope}[xshift=5cm]
          \draw[->] (0,0) -- (3.6,0);
          \draw[->] (0,0) -- (0,1.6);
          
          \draw (0,0) -- (0.8,0) -- (0.8,\h) -- (3.4,\h);
          
          \draw[fill=white] (0.8,-0.15) rectangle (2.2,-0.45);
          \node at (1.5,-0.3) {$a$};
        
          \node[right] at (3.5,\h) {$h$};
          \node at (1.7,1.3) {\textit{stepAtStart}$(a,h)$};
          
          \draw[dotted] (0,\h) -- (3.4,\h);
        \end{scope}
        
        \end{tikzpicture}
    \caption{Elementary cumulative functions.}
    \label{fig:pulse-step}
\end{figure}
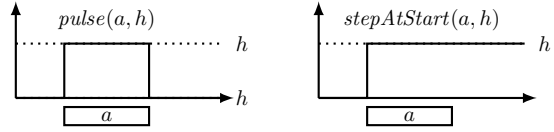

\textit{Cumulative functions} provide a powerful mechanism for modeling resource usage over time (see \cite{laborie2012interval,schaus2025implementing}). 
A cumulative function is a piecewise constant function $f: \mathbb{R} \rightarrow \mathbb{R}$ that represents the aggregated contribution of activities to a resource level at each time point.
Cumulative functions are constructed by combining \textit{elementary cumulative function expressions}, which define the contribution of individual interval variables or fixed time intervals. The main elementary cumulative functions, illustrated in Figure~\ref{fig:pulse-step}, are:
\begin{itemize}
    \item \textbf{pulse}$(a, h)$: Represents the contribution of interval $a$ with height $h$ during its execution. 
    If $a$ is present with $a = [s_a, e_a)$, the function equals $h$ for all $t \in [s_a, e_a)$ and 0 otherwise.    
    \item \textbf{step}$(t_0, h)$: Represents a constant contribution of height $h$ starting at time $t_0$ and continuing indefinitely. 
    For all $t \geq t_0$, the function equals $h$; otherwise it equals 0.
    \item \textbf{stepAtStart}$(a, h)$: Represents a step function that changes by $h$ at the start time of interval $a$. 
    If $a$ is present with $a = [s_a, e_a)$, the function equals $h$ for all $t \geq s_a$ and 0 otherwise.
\end{itemize}

Cumulative functions can be formed by combining functions using addition and subtraction: $f = \sum_{i} \epsilon_i \cdot f_i$ where $\epsilon_i \in \{-1, +1\}$ and each $f_i$ is also a cumulative function.
When all interval variables involved in a cumulative function are fixed, the function itself becomes a fixed piecewise constant function.
The constraint $ 0\leq f \leq C$ (or $f \geq C$) ensures that the cumulative function $f$ does not exceed (or fall below) capacity $C$ at any time point.

\subsection{Variables}
\label{sec:cpv}

Each task $i \in \mathcal{T}$ is modeled with an interval variable $a_i \in \mathcal{A}$.
The start and end are initialized to $s_i = [0, \Theta - d_i]$ and $e_i = [d_i, \Theta]$ respectively where
the horizon $\Theta$ can be set to the sum of all tasks duration.
These interval variables are always present.

The assignation of technicians to tasks is also represented by interval variables.
For each task $i \in \mathcal{T}$, for all technicians $j \in \mathcal{R}$, an optional interval variable $\omega_{ij}$ represents the possible assignment of technician $j$ to task $i$. 
The initial domain of these interval variables corresponds to the whole planning horizon ($[0, \Theta]$).
Task interval variables are always set to present, while the assignment interval variables are optional.
The unavailabilities of the technicians are also modeled as interval variables $\upsilon_{ju}$ that are set to the time windows corresponding to the unavailabilities.

\subsection{Constraints}
\label{sec:cpc}

The model is written as:
{\allowdisplaybreaks
\begin{align}
    & \text{minimize } \max\limits_{a_i \in \mathcal{A}}(e_i) \label{cp:obj} \\
    & \text{subject to} \nonumber \\
    & \textit{alternative}(a_i, \{ \omega_{ij} \mid j \in \mathcal{R} \}, \tau_i) \hspace{.11\linewidth} (i \in \mathcal{T}) \label{cp:alt} \\
    & \begin{multlined}[b][.89\linewidth]
        \textit{noOverlap}(\{ \omega_{ij} \mid i \in \mathcal{T} \} \cup \{ \upsilon_{ju} \mid u \in \mathcal{U}_j \}) \\
        (j \in \mathcal{R})
    \end{multlined} \label{cp:nover} \\
    & e_p \leq s_i \hspace{.45\linewidth} (i \in \mathcal{T}, p \in \mathcal{P}_i) \label{cp:prec} \\
    & \begin{multlined}[b][.89\linewidth]
        \textit{alternative}(a_i, \{ \omega_{ij} \mid j \in \mathcal{R} \wedge c_{iq} \in \mathcal{C}_j\}, \\
        \shoveleft[41pt]{x_{iq})} \hspace{.28\linewidth} (i \in \mathcal{T}, q \in \mathcal{Q}_i)
    \end{multlined} \label{cp:req} \\
    & o_l = \sum\limits_{a_i \in \mathcal{A} | l_i = l} pulse(a_i, \tau_i) \hspace{.24\linewidth} (l \in \mathcal{L}) \label{cp:occf}\\
    & 0 \leq o_l \leq k_l \hspace{.53\linewidth} (l \in \mathcal{L}) \label{cp:occ} \\
    & \begin{aligned}[b]
        b^{af} &= step(0, B^{af}) \\
        &+ \sum\limits_{\mathclap{a_i \in \mathcal{A} | z_{l_i} = \texttt{Aft}}} stepAtStart(a_i, m_i) \\
        &+ \sum\limits_{\mathclap{a_i \in \mathcal{A} | z_{l_i} = \texttt{Fwd}}} stepAtStart(a_i, -m_i)
    \end{aligned} \label{cp:baff} \\
    & \begin{aligned}[b]
        b^{lr} &= step(0, B^{lr}) \\
        &+ \sum\limits_{\mathclap{a_i \in \mathcal{A} | z_{l_i} = \texttt{Left}}} stepAtStart(a_i, m_i) \\
        &+ \sum\limits_{\mathclap{a_i \in \mathcal{A} | z_{l_i} = \texttt{Right}}} stepAtStart(a_i, -m_i)
    \end{aligned} \label{cp:blrf} \\
    & 0 \leq b^{af} \leq B^{af} \cdot 2 \label{cp:baf} \\
    & 0 \leq b^{lr} \leq B^{lr} \cdot 2 \label{cp:blr}
\end{align}
}

Constraints~\eqref{cp:alt} ensure that exactly $\tau_i$ technicians are selected for each task $i \in \mathcal{T}$.
Constraints~\eqref{cp:nover} ensure that each technician is assigned to a single task at the same time by enforcing that no overlap occurs between optional intervals of the technician.
These constraints also ensure that technicians are not assigned during their non-availability periods by also considering unavailability intervals $v_{ju}$ in the set of intervals that must not overlap.
Precedence constraints ensure that preceding activities are finished when an activity starts \eqref{cp:prec}.

\paragraph{Skill Requirements}
The skill requirements are enforced by constraints~\eqref{cp:req}.
They ensure that the number of technicians that possess a skill needed for a task and are assigned to the task is higher than or equal to the number required with this skill.
The variable $x_{iq}$ is a cardinality variable whose initial domain is $[n_{iq}, |\mathcal{R}_{iq}|]$ where $\mathcal{R}_{iq} = \{ j \in \mathcal{R} \mid c_{iq} \in \mathcal{C}_j\}$ is the set of all technicians that posses the skill required by the requirement $q$ of the task $i$.
Using a cardinality variable is necessary in order to allow more than the number of technicians with the skill required to be assigned to the task.

\paragraph{Locations Capacity}
Occupancy and balance constraints are modelled using cumulative functions.
For occupancy constraints, each location in the aircraft $l \in \mathcal{L}$ is modelled by a cumulative function $o_l$ \eqref{cp:occf} that tracks the number of technicians working in this location.
This cumulative function is linked to the task activities taking place at this location and must not overstep the capacity of the location $k_l$~\eqref{cp:occ}.

\paragraph{Balance}
For balance constraints, two cumulative functions are used:
The function $b^{af}$ \eqref{cp:baff} models the difference of mass between the front and rear of the aircraft (the \texttt{Aft} and \texttt{Fwd} zones).
The function $b^{lr}$ \eqref{cp:blrf} does the same for the left and right wings (the \texttt{Left} and \texttt{Right} zones).
When some weight is removed in one of these balance zones, it is either added to or subtracted from the relevant cumulative function.

For example, if an operation removes 30 units of weight on the left wing of the aircraft, this amount will be added to the cumulative function $b^{lr}$ while an operation that removes weight on the right wing will have this weight subtracted from the $b^{lr}$ function.
We use \textit{stepAtStart} functions to add these weights to the balance cumulative functions at the start time of the tasks.
In order to avoid having to deal with negative cumulative functions, these are shifted by the amount of tolerated mass difference ($B^{af}$ or $B^{lr}$).
These cumulative functions thus start at the tolerated mass difference ($B^{af}$ or $B^{lr}$) and must at all time be included between zero and twice the tolerated mass difference value, as ensured by constraints \eqref{cp:baf} and \eqref{cp:blr}.

\section{MIP Model}
\label{sec:mip}

This section presents the MIP model proposed to solve the problem.
Mixed Integer Programming (MIP) is a commonly used approach to model and solve combinatorial optimization problems. 
As mentioned in Section \ref{sec:rw}, MIP approaches have been successfully applied on similar problems.

The MIP model proposed to solve the Aircraft Dismantling Problem is based on the On/Off Event-based MIP Formulation (OOE) for the RCPSP from \cite{kone2011event} and \cite{artigues2013note}. The model has been modified by introducing additional balance constraints as well as resource assignment variables to capture technicians' skills and unavailabilities constraints.

This formulation is independent of the time span of the project and instead scales with the number of tasks.
The intuition is to discretize the time into events, which correspond to the starts of tasks.
Constraints that must be enforced at any time can then be decomposed and enforced at each event.

In this model, technicians non-availabilities are modeled as additional tasks with fixed time windows to which the unavailable technicians are assigned.
This set of additional tasks is denoted by $\mathcal{U}$.
The set of all tasks is thus $\mathcal{T}' = \mathcal{T} \cup \mathcal{U}$.
Let $\mathcal{E} = \{0, 1, \ldots, |\mathcal{T}’| - 1\}$ denote the index set of events $e$, 
which correspond to starts of a tasks.
We further define $\mathcal{E}^0 = \mathcal{E} \setminus \{0\}$.

Variables are introduced in Subsection \ref{sec:mipv} while
Subsection \ref{sec:mipc} details the constraints.
Finally, Subsection \ref{sec:ooe} explains how the model from \cite{kone2011event} was adapted for the specific constraints of this problem and Subsection \ref{sec:disc} discusses its complexity.

\subsection{Variables}
\label{sec:mipv}

The binary variables $z_{ie}$ for $i \in \mathcal{T}', e \in \mathcal{E}$ indicate if task $i$ is processed on occurence of event $e$:
\begin{multline}
    z_{ie} = \begin{cases}
        1 & \text{if task $i$ is processed at $e$} \\
        0 & \text{otherwise}
    \end{cases} \\
    (i \in \mathcal{T}'; e \in \mathcal{E})
\end{multline}
Additionally, auxiliary variables $z_{i,-1} = 0$ for $i \in \mathcal{T}'$ are used to deal with the case when $e = -1$ in some constraints.

The continuous variables $t_e$ for $e \in \mathcal{E}$ represent the time at which events occur.
The continuous variable $t_{max}$ corresponds to the makespan objective.

The binary variables $x_{ij}$ for $i \in \mathcal{T}', j \in \mathcal{R}$ correspond to the assignment of technician $j$ to task $i$:
\begin{multline}
    x_{ij} = \begin{cases}
        1 & \text{if technician $j$ is assigned to task $i$} \\
        0 & \text{otherwise}
    \end{cases} \\
    (i \in \mathcal{T}'; j \in \mathcal{R})
\end{multline}

The binary variables $y_{ije}$ for $i \in \mathcal{T}', j \in \mathcal{R}, e \in \mathcal{E}$ are event assignment variables:
\begin{multline}
    y_{ije} = \begin{cases}
        1 & \text{if technician $j$ is assigned to task $i$} \\
        & \text{during event $e$} \\
        0 & \text{otherwise}
    \end{cases} \\
    (i \in \mathcal{T}'; j \in \mathcal{R}, e \in \mathcal{E})
\end{multline}
These variables are used in some constraints and make the link between the variables $z_{ie}$ and $x_{ij}$.

The binary variables $a_{ie}$ for $i \in \mathcal{T}', e \in \mathcal{E}$ indicate if task $i$ starts at event $e$:
\begin{multline}
    a_{ie} = \begin{cases}
        1 & \text{if task $i$ starts at event $e$} \\
        0 & \text{otherwise}
    \end{cases} \\
    (i \in \mathcal{T}'; e \in \mathcal{E})
\end{multline}

Finally, the continuous variables $b^{af}_{e}$ and $b^{lr}_{e}$ for $e \in \mathcal{E}$ track the balance of the aft-forward and left-right axis respectively and are used for balance constraints.
Two auxiliary variable $b^{af}_{-1} = 0$ and $b^{lr}_{-1} = 0$ are used for the case $e = -1$.

\subsection{Constraints}
\label{sec:mipc}

The MIP model is written as:
{\allowdisplaybreaks
\begin{align}
    & \text{minimize } t_{max} \label{mip:obj} \\
    & \text{subject to} \nonumber \\
    & t_0 = 0 \label{mip:t0} \\
    & t_e - t_{e-1} \geq 0 \hspace{.44\linewidth} (e \in \mathcal{E}^0)  \label{mip:tpos} \\
    & z_{i,-1} = 0 \hspace{.53\linewidth} (i \in \mathcal{T}') \label{mip:z0} \\
    & a_{ie} \geq z_{ie} - z_{i,e-1} \hspace{.24\linewidth} (i \in \mathcal{T}'; e \in \mathcal{E}) \label{mip:aie1} \\
    & a_{ie} \leq 1 - z_{i,e-1} \hspace{.27\linewidth} (i \in \mathcal{T}'; e \in \mathcal{E}) \label{mip:aie2} \\
    & \sum_{e \in \mathcal{E}} z_{ie} \geq 1 \hspace{.49\linewidth} (i \in \mathcal{T}') \label{mip:wmin} \\
	& \begin{multlined}[b][.85\linewidth]
        \sum_{e'=0}^{e-1} z_{ie'} - e(1 - (z_{ie} - z_{i,e-1})) \leq 0 \\
        (i \in \mathcal{T}'; e \in \mathcal{E}^0)
    \end{multlined} \label{mip:zcont1} \\
    & \begin{multlined}[b][.85\linewidth]
        \sum_{e'=e}^{N-1} z_{ie'} - (N - e)(1 + (z_{ie} - z_{i,e-1})) \leq 0 \\
         (i \in \mathcal{T}'; e \in \mathcal{E}^0)
    \end{multlined} \label{mip:zcont2} \\
    & \begin{multlined}[b][.85\linewidth]
        t_f - t_e - d_i ( z_{ie} - z_{i,e-1} - (z_{if} - z_{i,f-1}) ) \geq -d_i \\
        \hspace{.1\linewidth} (i \in \mathcal{T}'; e, f \in \mathcal{E} \mid e < f)
    \end{multlined} \label{mip:tzlink} \\
    & \begin{multlined}[b][.85\linewidth]
        t_{max} \geq t_e + d_i - \Theta ( 1 - a_{ie}) \\
        (i \in \mathcal{T}, e \in \mathcal{E})
    \end{multlined} \label{mip:mksp} \\
    & y_{ije} \leq x_{ij} \hspace{.25\linewidth} (i \in \mathcal{T}'; j \in \mathcal{R}; e \in \mathcal{E}) \label{mip:xyzlink1} \\
    & y_{ije} \leq z_{ie} \hspace{.25\linewidth} (i \in \mathcal{T}'; j \in \mathcal{R}; e \in \mathcal{E}) \label{mip:xyzlink2} \\
    & y_{ije} \geq x_{ij} + z_{ie} - 1 \hspace{.06\linewidth} (i \in \mathcal{T}'; j \in \mathcal{R}; e \in \mathcal{E}) \label{mip:xyzlink3} \\
    & \sum_{j \in \mathcal{R}} y_{ije} = \tau_i z_{ie} \hspace{.27\linewidth} (i \in \mathcal{T}'; e \in \mathcal{E}) \label{mip:yzlink} \\
    & \sum_{i \in \mathcal{T}'} y_{ije} \leq 1 \hspace{.33\linewidth} (j \in \mathcal{R}; e \in \mathcal{E}) \label{mip:nooverlap} \\
    & x_{ij} \leq \sum_{e \in \mathcal{E}} y_{ije} \leq N x_{ij} \hspace{.14\linewidth} (i \in \mathcal{T}'; j \in \mathcal{R}) \label{mip:xylink} \\
    & \begin{multlined}[b][.87\linewidth]
        1 - x_{ij} \leq \sum_{e \in \mathcal{E}} z_{ie} - \sum_{e \in \mathcal{E}} y_{ije} \leq N (1 - x_{ij}) \\
        (i \in \mathcal{T}'; j \in \mathcal{R})
    \end{multlined} \label{mip:wylink} \\
    & \sum_{j \in \mathcal{R}} x_{ij} = \tau_i \hspace{.46\linewidth} (i \in \mathcal{T}') \label{mip:ass} \\
    & x_{uj} = 1 \hspace{.4\linewidth} (j \in \mathcal{R}, u \in \mathcal{U}_j) \label{mip:uset} \\
    & t_e \geq s_{ju} a_{ue} \hspace{.21\linewidth} (j \in \mathcal{R}, u \in \mathcal{U}_j, e \in \mathcal{E}) \label{mip:ustart} \\
    & \begin{multlined}[b][.87\linewidth]
        t_e \leq s_{ju} a_{ue} + \Theta (1 - a_{ue}) \\
        (j \in \mathcal{R}, u \in \mathcal{U}_j, e \in \mathcal{E})
    \end{multlined} \label{mip:uend} \\
    & \begin{multlined}[b][.87\linewidth]
        z_{pe} + \sum_{e'=0}^e z_{ie'} - e(1 - z_{pe}) \leq 1 \\
        (i \in \mathcal{T}; p \in \mathcal{P}_i; e \in \mathcal{E})
    \end{multlined} \label{mip:prec} \\
    & \sum_{j \in \mathcal{R} \mid c_{iq} \in \mathcal{C}_j} x_{ij} \geq n_{iq} \hspace{.17\linewidth} (i \in \mathcal{T}; q \in \mathcal{Q}_i) \label{mip:req} \\
    & \sum_{i \in \mathcal{T} \mid l_i = l} \tau_i z_{ie} \leq k_l \hspace{.25\linewidth} (l \in \mathcal{L}; e \in \mathcal{E}) \label{mip:occ} \\
    & \begin{multlined}[b][.87\linewidth]
        b^{af}_{e} = b^{af}_{e-1} + \sum\limits_{\mathclap{i \in \mathcal{T} | z_{l_i} = \texttt{Aft}}} a_{ie} m_i + \sum\limits_{\mathclap{i \in \mathcal{T} | z_{l_i} = \texttt{Fwd}}}  a_{ie} (-m_i) \\
        (e \in \mathcal{E})
    \end{multlined} \label{mip:baff} \\
    & \begin{multlined}[b][.87\linewidth]
        b^{lr}_{e} = b^{lr}_{e-1} + \sum\limits_{\mathclap{i \in \mathcal{T} | z_{l_i} = \texttt{Left}}} a_{ie} m_i + \sum\limits_{\mathclap{i \in \mathcal{T} | z_{l_i} = \texttt{Right}}}  a_{ie} (-m_i) \\
        (e \in \mathcal{E})
    \end{multlined} \label{mip:blrf} \\
    & -B^{af} \leq b^{af}_{e} \leq B^{af} \hspace{.32\linewidth} (e \in \mathcal{E}) \label{mip:baf} \\
    & -B^{lr} \leq b^{lr}_{e} \leq B^{lr} \hspace{.35\linewidth} (e \in \mathcal{E}) \label{mip:blr}
\end{align}
}

Constraints \eqref{mip:t0} and \eqref{mip:tpos} ensure that the timing of events follow an increasing order: $t_0 \leq t_1 \leq ... \leq t_{N-1}$.
Constraints \eqref{mip:z0} initialize $z_{ie}$ variables while constraints \eqref{mip:wmin} ensure that each task is processed during at least one event.
Constraints~\eqref{mip:aie1} and~\eqref{mip:aie2} link together the task start variable $a_{ie}$ with the task processing variables $z_{ie}$.
Constraints \eqref{mip:zcont1} and \eqref{mip:zcont2} make sure that each task is processed in a contiguous block of events:
Constraints \eqref{mip:zcont1} make sure that when a task is processed at an event $e$ but not the previous one ($z_{ie} - z_{i,e-1} = 1$), it is never processed before ($\sum_{e'=0}^{e-1} z_{ie'} \leq 0$).
Constraints \eqref{mip:zcont2} enforce that when a task is not processed at an event $e$ but processed at the previous one ($z_{ie} - z_{i,e-1} = -1$), it is never processed after ($\sum_{e'=e}^{N-1} z_{ie'} \leq 0$).

Constraints \eqref{mip:tzlink} enforce the processing time of tasks based on task processing variables $z_{ie}$ and time variables $t_e$.
They ensure that for any two events $e, f \in \mathcal{E}$ with $e < f$, task $i$ may start at event $e$ and end at event $f$ only if the difference of time between $f$ and $e$ is larger than or equal to the duration of task $i$.

Constraints \eqref{mip:mksp} ensure that the makespan objective \eqref{mip:obj} is not smaller than the completion time of any task.

Constraints \eqref{mip:xyzlink1} to \eqref{mip:xyzlink3} link together event assignment variables $y_{ije}$ with assignment variables $x_{ij}$ and task processing variables $z_{ie}$ following the relation $y_{ije} = x_{ij} \cdot z_{ie}$.
Constraints \eqref{mip:yzlink} link the task processing variables $z_{ie}$ with the event assignment variables $y_{ije}$ and ensure that the correct number of technicians is used when a task is processed.
Constraints \eqref{mip:nooverlap} prevent a technician to be concurrently assigned to more than one task.
Constraints \eqref{mip:xylink} link assignment variables $x_{ij}$ with event assignment variables $y_{ije}$.
Constraints \eqref{mip:wylink} make sure that the number of active event assignment variables $\sum y_{ije}$ is equal to either the number of active task processing events $\sum z_{ie}$ if the assignment variable $x_{ij}$ has a value of $1$ or $0$ if $x_{ij} = 0$.
Constraints \eqref{mip:ass} ensure that the correct number of technicians is assigned to each task.

Constraints \eqref{mip:uset} fix the unavailability tasks to the corresponding technician.
Constraints \eqref{mip:ustart} and \eqref{mip:uend} fix the events of the the unavailability tasks.
Constraints \eqref{mip:prec} set up precedences between tasks: if a precedence task $p$ is processed last at event $e$, then $\sum_{e'=0}^{e} z_{ie'} = 0$ which implies that task $i$ must be processed later.

\paragraph{Skill Requirements}
Constraints \eqref{mip:req} ensure that requirements are respected:
For each requirement $q \in \mathcal{Q}_i$ of each task $i \in \mathcal{T}$, a constraint ensures that the sum of technicians assigned to the task that have the requested skill ($j \in \mathcal{R} \mid c_{iq} \in \mathcal{C}_j$) is greater than or equal to the requested amount $n_{iq}$.

\paragraph{Locations Capacity}
Constraints \eqref{mip:occ} model the locations capacity constraint of the problem by limiting the cumulative occupation of tasks for each location $l \in \mathcal{L}$ during each event.

\paragraph{Balance}
Constraints \eqref{mip:baff} and \eqref{mip:blrf} set up balance variables $b^{af}_e$ and $b^{lr}_e$: At each event $e$, the balance is equal to the balance at the previous event ($b^{af}_{e-1}$ or $b^{lr}_{e-1}$) plus the balance change at this event, which corresponds to the value of mass removed during tasks impacting the balance axis that start at event $e$, which is either added or removed depending on the location of the task:
\begin{equation}
    \sum\limits_{\mathclap{i \in \mathcal{T} | z_{l_i} = \texttt{Aft/Left}}} a_{ie} m_i \qquad + \sum\limits_{\mathclap{i \in \mathcal{T} | z_{l_i} = \texttt{Fwd/Right}}} a_{ie}(-m_i)
\end{equation}

Finally, constraints \eqref{mip:baf} and \eqref{mip:blr} enforce that the balance variables $b^{af}_e$ and $b^{lr}_e$ stay within the balance range at each event $e$.

\subsection{Differences from the OOE Model}
\label{sec:ooe}

A key difference from the model of \cite{kone2011event} is the introduction of the resource assignment variables $x_{ij}$ and $y_{ije}$, which capture additional constraints related to technicians’ skills and unavailabilities.

The inclusion of technician unavailability tasks, modeled as fixed activities that do not contribute to the makespan, also required a careful adaptation of the formulation proposed by \cite{kone2011event}.
In their model, the makespan constraints can be expressed as
$$
t_{\max} \geq t_e + d_i (z_{ie} - z_{i,e-1}) \quad (i \in \mathcal{T},\, e \in \mathcal{E}),
$$
but this expression is not directly applicable in our case, since unavailability tasks are also associated with events and time intervals.

Regarding the balancing constraint, it might be tempting—but incorrect—to formulate it as
$$
 - B^{af} \leq 
 \sum\limits_{\mathclap{i \in \mathcal{T} \mid z_{l_i} = \texttt{Aft}}} z_{ie} m_i +
 \sum\limits_{\mathclap{i \in \mathcal{T} \mid z_{l_i} = \texttt{Fwd}}} z_{ie} (-m_i)
 \leq B^{af} \quad (e \in \mathcal{E}),
$$
since, in the case of activities with identical start times, the model could artificially choose event assignments that satisfy the balance constraints without reflecting a valid configuration.
In our formulation, the weight differences are accumulated across start events through the variables $a_{ie}$, ensuring that the model cannot “cheat” regardless of the event assignments.
If a feasible solution exists, then there is at least one corresponding event assignment for which the balance constraints are satisfied.

\subsection{Discussion}
\label{sec:disc}

While this MIP model avoids a decomposition in time steps, which scales poorly on larger problems, it remains costly in terms of variables and constraints with $O(|\mathcal{T}'|^2|\mathcal{R}|)$ binary variables, $O(|\mathcal{T}'|)$ continuous variables and $O(|\mathcal{T}'|^2|\mathcal{R}| + (|\mathcal{P}| + |\mathcal{R}|)|\mathcal{T}'|)$ constraints where $\mathcal{P} = \cup_{i \in \mathcal{T}} \mathcal{P}_i$ is the union of the sets of precedences of each task $i \in \mathcal{T}$.
Furthermore, the decomposition in events introduces symmetries if several tasks are started at the same time.
In this case, several events will share the same time and may be interchangeable.

\section{Experiments}
\label{sec:xp}

This section presents the experiments done with both models and their results.
Two experiments are considered:
The first one compares both approaches on the set of all instances.
The second one compares the anytime behavior of the CP approach on several variations of the problem where specific constraints are deactivated to examine their impact on the problem.

Subsection \ref{sec:data} explains how industrial data was used to generate the instances used for the experiments.
The experimental setup for the two experiments is detailed in Subsection \ref{sec:setup}.
Finally, Subsection \ref{sec:res} presents the results of the two experiments.

\subsection{Data}
\label{sec:data}

The instances used in the experiments are based on real industrial data provided by our partner within the Planum research project. The dataset was collected during the complete dismantling of a Boeing 737NG-600 aircraft, and it describes 1454 disassembly tasks representing the full set of operations performed during the process.

Constructing this dataset required a substantial collaborative effort. Since no structured numerical data were available, the information had to be manually reconstructed from technical documentation and user manuals to extract task precedences, durations, and other relevant attributes. In addition, a detailed labeling of operations was carried out during the actual dismantling to ensure that the model accurately reflects the real industrial process.

Most of the task data originate from this source, with the exception of the mass values used for the balance constraints. Because the industrial partner is still in the process of collecting precise mass information, these values were approximated artificially: a mass between 5 kg and 500 kg was assigned to 214 tasks located in the four balance zones, while the remaining tasks were considered to have no mass impact. 
The maximum allowable mass difference was set to 300 kg on the Aft–Forward axis and 500 kg on the Left–Right axis.

Two different skills are considered, which correspond to certifications needed in the air transport industry: B1 and B2.
Task durations are calculated based on the industrial data.
A unit of time corresponds to 15 minutes, which is the smallest time precision observed in the real world data.
There are 13 different locations with capacities varying between 2 and 10.
An additional dummy location with an infinite capacity is used for a few tasks that either apply to the whole plane or for which the location is not relevant.

The instances used in the experiments were created based on this dataset.
The instance B737NG600-1454 corresponds to the whole set of tasks.
15 smaller instances of various sizes were generated based on this instance by removing some of the tasks.
The tasks removed are selected randomly.
The instances are named with the following convention: \texttt{B737NG600-<number of tasks>.json}.

Each instance uses the same set of technicians with 7 technicians available.
Among them, one has the B1 certification, one has the B2 certification, one has both B1 and B2 certifications and the four others have no certification.
Some unavailability periods are randomly assigned to the technicians.

\begin{figure*}
    \centering
    \includegraphics[width=1\linewidth]{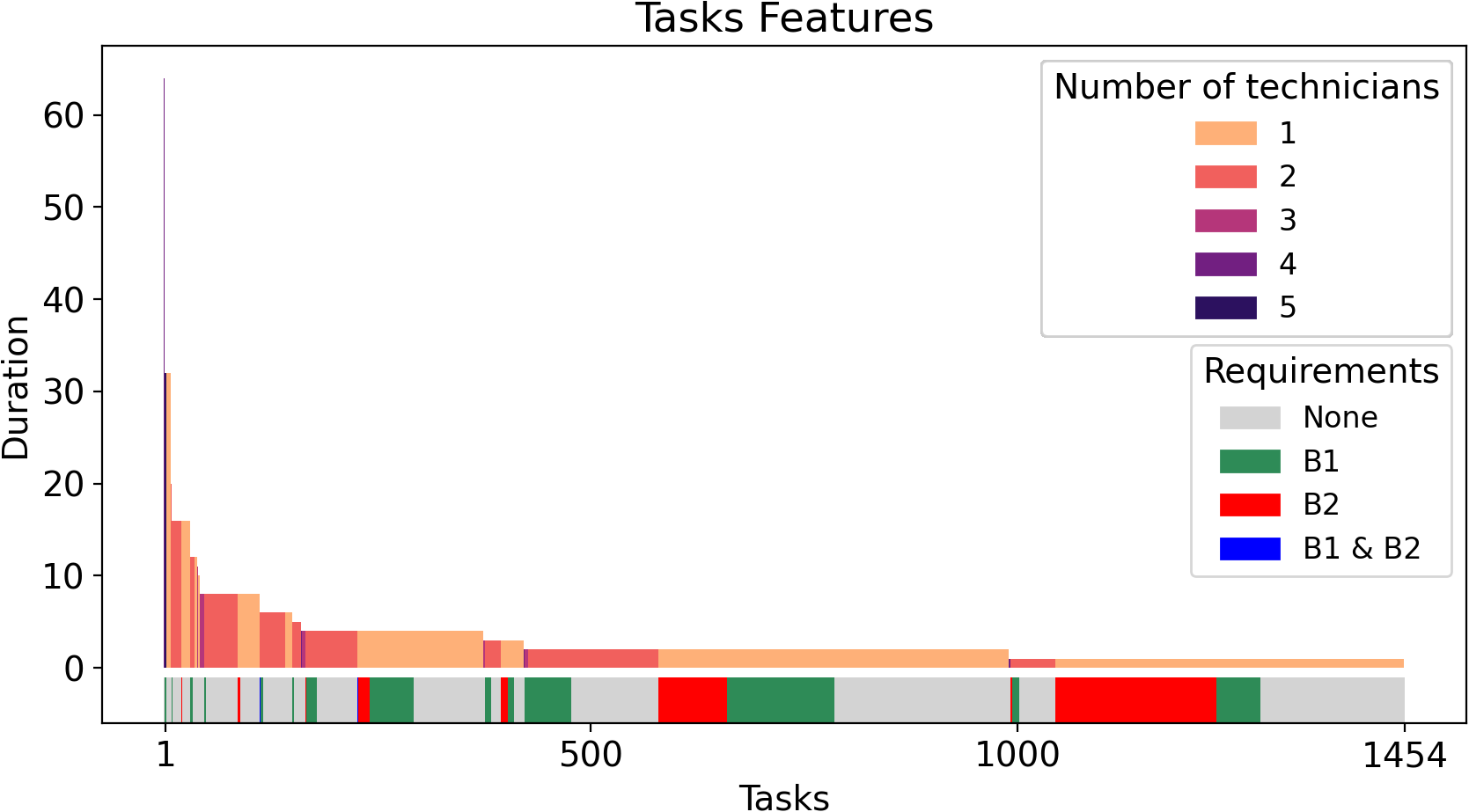}
    \caption{Durations and number of technicians for the tasks of instance B737NG600-1454}
    \label{fig:histogram}
\end{figure*}

Figure \ref{fig:histogram} shows a visualization of some of the tasks features of the instance B737NG600-1454.
Tasks are sorted by duration then grouped by number of technicians required and skill requirements.
The height of the bars in the top part of the graph show the duration of each task.
The colors in the top part show the number of technicians required.
The colors in the bottom part show the skill requirements.

We can see that the vast majority of tasks have a duration under 10 and require no more than one or two technicians.
There is one single longest task that has a duration of 64 and requires 4 technicians.
There is approximately $22\%$ of tasks with a B1 requirement, approx. $20\%$ with a B2 requirement and 2 tasks with both B1 and B2.
The remaining tasks have no requirement.

An anonymized version of the instances as well as the models and results is made available at the following repository: \url{https://github.com/cftmthomas/AircraftDisassembly}.

\subsection{Experimental setup}
\label{sec:setup}

We compare the CP and MIP approaches on the set of all instances.
Additionally, we compare the performances of the CP approach on the base problem with all its constraints as well as several relaxations where some of the constraints have been deactivated in order to examine their impact on the problem difficulty:

\paragraph{Requirements relaxation}
The first relaxation consists in discarding the certifications requirements for all tasks.
For the CP model, this consists in removing constraints \eqref{cp:req}.

\paragraph{Capacity relaxation}
This relaxation consists in deactivating the locations capacities constraints.
Constraints \eqref{cp:occf} and \eqref{cp:occ} are deactivated in the CP model.

\paragraph{Balance relaxation}
The third relaxation consists in deactivating the balance constraints for both balance axis.
For the CP model, this consist in removing constraints \eqref{cp:baff}, \eqref{cp:blrf}, \eqref{cp:baf} and \eqref{cp:blr}.

\paragraph{Requirements, capacity and balance relaxation}
The last relaxation considered combines the three previous ones by discarding the requirements, locations capacity and balance constraints.
It is the closest to a classical RCPSP with the only difference being that technicians can be unavailable during some parts of the scheduling period.

The CP Model is implemented in CP Optimizer with the default search used.
This search consists in an Adpative LNS (ALNS) approach \citep{laborie2018ibm} enhanced with a failure directed search to prove optimality.
The MIP model is implemented with CPLEX mixed integer optimizer with its default search used for the experiments.
Both solvers are part of IBM's CPLEX optimization suite whose version 22.1.1 was used in the experiments.

Experiments were conducted on a Linux server with 2 processors Intel(R) Xeon(R) E5-2687W (40 threads total) and 128 GB of RAM.
Both approaches were limited to a single thread for each run with a time limit (denoted by $tmt$) of 3600 seconds and a maximum memory limit of 8GB.

\subsubsection{Performance measurement}
\label{sec:perf}

We compare the results using the primal integral method proposed in \cite{Berthold2013}.
Intuitively, this performance metric consists in measuring the area under the anytime objective curve during the whole search.
In order to compare performances over different instances, the primal integral is computed based on the primal gap, which normalizes an objective value based on the optimum or best known objective.
Given a solution $x$, an optimal (or best known) solution $x^*$ and an objective function $o()$, the primal gap $\gamma \in [0, 1]$ is defined as:
\begin{equation}
    \gamma(x) = \begin{cases}
        0 & \text{if } o(x^*) = o(x) = 0 \\
        1 & \text{if } o(x^*) \cdot o(x) < 0 \\
        \frac{|o(x^*) - o(x)|}{max(|o(x^*)|, |o(x)|)} & \text{otherwise}
    \end{cases}
    \label{eq:pgap}
\end{equation}
The primal gap thus corresponds to the ratio between the distance of the current objective to the best objective $|o(x^*) - o(x)|$ and the largest absolute value between the two.
This ratio tends to 1 if the current objective tends to $\infty$ and reaches 0 if $|o(x^*)| = |o(x)|$.

For an experimental run where several improving solutions have been found at given points in time, the primal gap can be used to compute a primal gap step function $p(t)$, which is defined as:
\begin{equation}
    p(t) = \begin{cases}
        1 & \text{if no incumbent until } t \\
        \gamma(x(t)) & \text{otherwise}
    \end{cases}
    \label{eq:pfun}
\end{equation}
where $x(t)$ is the incumbent solution at time $t$.
The function $p(t)$ starts at 1 until a first solution has been found and decreases to reach 0 once the optimum or best known solution has been found.

The primal integral $P(T)$ for a time $T \in [0, t_{max}]$ is the integral of the primal gap function from $0$ to $T$:
\begin{equation}
    P(T) = \int_{t=0}^{T} p(t) dt = \sum_{i=1}^{I} p(t_{i-1}) \cdot (t_i-t_{i-1})
    \label{eq:pint}
\end{equation}
where $t_0=0$, $t_i \in [0,T]$ for $i \in 1, \dots, I$ denotes the time points at which a solution has been found and $t_I = T$.

\subsection{Computational results}
\label{sec:res}

\paragraph{Models comparison}
This first experiment consists in comparing both approaches on all instances for the full problem with all its constraints.

\begin{table*}
    \centering
    \caption{Results on the problem with all constraints}
    \begin{tblr}{
      row{1} = {c},
      row{2} = {c},
      cell{1}{1} = {c=2}{},
      cell{1}{3} = {c=3}{},
      cell{1}{6} = {c=3}{},
      cell{3}{2} = {r},
      cell{3}{3} = {r},
      cell{3}{4} = {r},
      cell{3}{5} = {r},
      cell{3}{6} = {r},
      cell{3}{7} = {r},
      cell{3}{8} = {r},
      cell{4}{2} = {r},
      cell{4}{3} = {r},
      cell{4}{4} = {r},
      cell{4}{5} = {r},
      cell{4}{6} = {r},
      cell{4}{7} = {r},
      cell{4}{8} = {r},
      cell{5}{2} = {r},
      cell{5}{3} = {r},
      cell{5}{4} = {r},
      cell{5}{5} = {r},
      cell{5}{6} = {r},
      cell{5}{7} = {r},
      cell{5}{8} = {r},
      cell{6}{2} = {r},
      cell{6}{3} = {r},
      cell{6}{4} = {r},
      cell{6}{5} = {r},
      cell{6}{6} = {r},
      cell{6}{7} = {r},
      cell{6}{8} = {r},
      cell{7}{2} = {r},
      cell{7}{3} = {r},
      cell{7}{4} = {r},
      cell{7}{5} = {r},
      cell{7}{6} = {r},
      cell{7}{7} = {r},
      cell{7}{8} = {r},
      cell{8}{2} = {r},
      cell{8}{3} = {r},
      cell{8}{4} = {r},
      cell{8}{5} = {r},
      cell{8}{6} = {r},
      cell{8}{7} = {r},
      cell{8}{8} = {r},
      cell{9}{2} = {r},
      cell{9}{3} = {r},
      cell{9}{4} = {r},
      cell{9}{5} = {r},
      cell{9}{6} = {r},
      cell{9}{7} = {r},
      cell{9}{8} = {r},
      cell{10}{2} = {r},
      cell{10}{3} = {r},
      cell{10}{4} = {r},
      cell{10}{5} = {r},
      cell{10}{6} = {r},
      cell{10}{7} = {r},
      cell{10}{8} = {r},
      cell{11}{2} = {r},
      cell{11}{3} = {r},
      cell{11}{4} = {r},
      cell{11}{5} = {r},
      cell{11}{6} = {r},
      cell{11}{7} = {r},
      cell{11}{8} = {r},
      cell{12}{2} = {r},
      cell{12}{3} = {r},
      cell{12}{4} = {r},
      cell{12}{5} = {r},
      cell{12}{6} = {r},
      cell{12}{7} = {r},
      cell{12}{8} = {r},
      cell{13}{2} = {r},
      cell{13}{3} = {r},
      cell{13}{4} = {r},
      cell{13}{5} = {r},
      cell{13}{6} = {r},
      cell{13}{7} = {r},
      cell{13}{8} = {r},
      cell{14}{2} = {r},
      cell{14}{3} = {r},
      cell{14}{4} = {r},
      cell{14}{5} = {r},
      cell{14}{6} = {r},
      cell{14}{7} = {r},
      cell{14}{8} = {r},
      cell{15}{2} = {r},
      cell{15}{3} = {r},
      cell{15}{4} = {r},
      cell{15}{5} = {r},
      cell{15}{6} = {r},
      cell{15}{7} = {r},
      cell{15}{8} = {r},
      cell{16}{2} = {r},
      cell{16}{3} = {r},
      cell{16}{4} = {r},
      cell{16}{5} = {r},
      cell{16}{6} = {r},
      cell{16}{7} = {r},
      cell{16}{8} = {r},
      cell{17}{2} = {r},
      cell{17}{3} = {r},
      cell{17}{4} = {r},
      cell{17}{5} = {r},
      cell{17}{6} = {r},
      cell{17}{7} = {r},
      cell{17}{8} = {r},
      cell{18}{2} = {r},
      cell{18}{3} = {r},
      cell{18}{4} = {r},
      cell{18}{5} = {r},
      cell{18}{6} = {r},
      cell{18}{7} = {r},
      cell{18}{8} = {r},
      vline{1,3,6,9} = {1-18}{},
      hline{1,3,19} = {-}{},
    }
    Instance       &        & CP           &       &       & MIP          &       &         \\
    Name           & $obj^*$ & $P(tmt)$ & $obj$ & $t^*$  & $P(tmt)$ & $obj$ & $t^*$    \\
    B737NG600-10   & 64     & 0.022        & 64    & 0.044 & 58.633       & 64    & 179.661 \\
    B737NG600-15   & 64     & 0.007        & 64    & 0.009 & 238.449      & 64    & 463.419 \\
    B737NG600-20   & 65     & 0.023        & 65    & 0.048 & 1480.616     & 72    & -       \\
    B737NG600-30   & 68     & 0.043        & 68    & 1.914 & 2453.189     & 128   & -       \\
    B737NG600-40   & 91     & 0.057        & 91    & -     & -            & -     & -       \\
    B737NG600-50   & 93     & 0.108        & 93    & -     & -            & -     & -       \\
    B737NG600-75   & 114    & 0.114        & 114   & -     & -            & -     & -       \\
    B737NG600-100  & 117    & 0.152        & 117   & -     & -            & -     & -       \\
    B737NG600-150  & 159    & 0.205        & 159   & -     & -            & -     & -       \\
    B737NG600-200  & 184    & 0.410        & 184   & -     & -            & -     & -       \\
    B737NG600-300  & 250    & 1.602        & 250   & -     & -            & -     & -       \\
    B737NG600-400  & 287    & 1.102        & 287   & -     & -            & -     & -       \\
    B737NG600-600  & 420    & 7.581        & 420   & -     & x            & x     & x       \\
    B737NG600-800  & 505    & 15.697       & 505   & -     & x            & x     & x       \\
    B737NG600-1200 & 834    & 17.789       & 834   & -     & x            & x     & x       \\
    B737NG600-1454 & 973    & 31.022       & 973   & -     & x            & x     & x       
    \end{tblr}
    \label{tab:resall}
\end{table*}

Table \ref{tab:resall} presents the results for the problem with all constraints.
The columns \textit{name} and $obj^*$ indicate the name and best known solution for each instance.
For each approach, the column $P(tmt)$ shows the primal integral (computed with $tmt = 3600$); the column $obj$ shows the best solution obtained and the column $t^*$ indicates the time to prove optimality if the approach was able to.
The character "\texttt{-}" indicates a timeout and the character "\texttt{x}" indicates that the approach ran out of memory.

As we can see, the CP approach vastly outperforms the MIP approach, which is only able to solve small sized instances, up to 30 tasks.
The CP approach is able to prove optimality for instances up to 30 tasks but stagnates until timeout for the larger instances.

The MIP approach is only able to find a solution on the four smallest instances and to solve only two of them to optimality.
Furthermore, it runs out of memory (8G per thread) on instances of 600 or more tasks.
This can be explained by the fact that the MIP model uses a number of constraints and variables that grow quadratically with the number of tasks.
Additional experiments have shown that this behavior stays the same on all variations of the problem and in some cases is even worse.
The full results of these experiments are available in the repository \url{https://github.com/cftmthomas/AircraftDisassembly}.
Due to its poor performances, the MIP model is not considered in the next experiment.

\paragraph{Anytime performances of the CP approach}

This experiment aims at comparing the impact of different constraints on the problem difficulty.
To do so, we compare the anytime behavior of the CP approach on several variations of the problem where some constraints are deactivated.

\begin{figure*}
    \centering
    \includegraphics[width=1\linewidth]{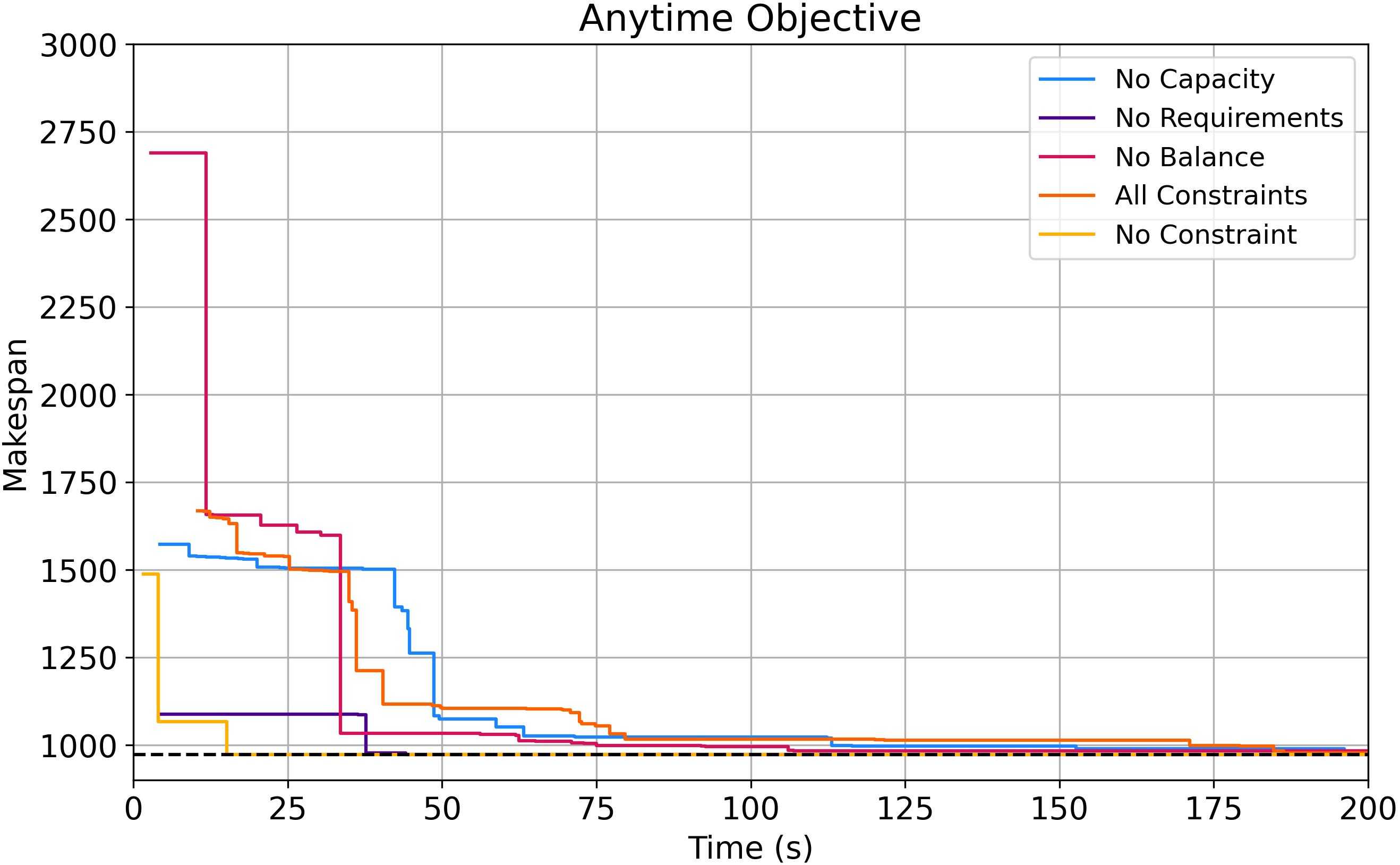}
    \caption{Anytime performances of the CP approach on the instance B373NG600-1454}
    \label{fig:anytime}
\end{figure*}
Figure \ref{fig:anytime} shows the anytime objective value of the CP approach on the largest instance B373NG600-1454 for the different variations considered.
We show only the start of the search, from 0 to 200s, as this is the part with the most variation in the objective values.
After that time, we observe mostly stagnation with occasional very small improvements of the objective.
Eventually, a best known solution is reached, which has the same objective value of 973 for all variations of the problem.
It is indicated by the horizontal dashed line.

This stagnation combined with the fact that all the variations have the same objective for their best known solution suggests that the three constraints considered (Requirements, Balance and Capacity) do not impact much the optimal makespan.
One possible explanation is the characteristics of the instance, which consists of many small tasks with short durations, each requiring only one or two technicians.
This provides a high degree of flexibility in accommodating these constraints, which may explain their lack of impact on the makespan.

We observe that deactivating constraints improves the anytime behavior, with the relaxation of the \emph{Requirements} constraint having the most significant impact. This can be attributed to the relatively large number of tasks with specific requirements, coupled with the fact that only three technicians possess the necessary certifications.

The results are less explicit when considering the Capacity or Balance constraint individually.
Removing these constraints seems to allow finding a first solution earlier compared to the variation with all constraints.
However the  anytime behavior on these relaxations does not clearly dominate the version with all constraints.
The lower impact of these constraints can again be explained by the large number of small tasks, which offer a lot of flexibility in the scheduling.

\section{Conclusion}
\label{sec:ccl}

We presented the aircraft disassembly scheduling problem.
This problem, which consists in scheduling a set of dismantling tasks while assigning human resources to them, is a variation of the RCPSP.
The problem also deals with skill requirements and additional constraints related to the balance and capacity in some parts of the aircraft.

We proposed two approaches to tackle the problem.
The first one is a constraint programming model using advanced modeling features including conditional task intervals, sequence variables and cumulative functions.
The second one is a MIP model.

Experiments were run on 16 instances derived from real data provided by an industrial partner.
The experiments show the CP model vastly outperforms the MIP model and is able to find feasible solutions for instances of up to 1450 tasks.
Several variations of the problem were considered where some constraints are removed to examine their impact on the problem difficulty.
The results indicate that the requirements constraint (some technicians assigned to some tasks require specific certifications) is the one that affects the most the anytime behavior on the problem.

\subsection{Future Work}

A possible future research avenue would be to investigate metaheuristic methods to solve the problem, as these methods have been successfully applied to variants of the RCPSP \citep{laurent2017new, mischek2021local}.
Including a possible uncertainty in the duration of the task could be useful, and the solutions can then be made more robust against this uncertainty using the techniques introduced in \cite{davenport2001slack}.

\appendix



\bibliography{references.bib}

@inproceedings{davenport2001slack,
  title={Slack-based techniques for robust schedules},
  author={Davenport, Andrew J and Gefflot, Christophe and Beck, J Christopher and others},
  booktitle={{Proceedings of the Sixth European Conference on Planning (ECP-2001)}},
  pages={7--18},
  year={2001}
}

@article{schaus2025implementing,
  title={Implementing Cumulative Functions with Generalized Cumulative Constraints},
  author={Schaus, Pierre and Thomas, Charles and Kameugne, Roger},
  journal={arXiv preprint arXiv:2508.01751},
  year={2025}
}

@phdthesis{de1998resource,
  title={Resource-constrained multi-project management},
  author={De Boer, Ronald},
  year={1998},
  school={{PhD thesis, University of Twente, The Netherlands}}
}

@article{neumann2003project,
  title={Project scheduling with inventory constraints},
  author={Neumann, Klaus and Schwindt, Christoph},
  journal={Mathematical Methods of Operations Research},
  volume={56},
  number={3},
  pages={513--533},
  year={2003},
  publisher={Springer}
}

@article{vilhelmsen2016heuristic,
  title={A heuristic and hybrid method for the tank allocation problem in maritime bulk shipping},
  author={Vilhelmsen, Charlotte and Larsen, Jesper and Lusby, Richard},
  journal={4OR},
  volume={14},
  number={4},
  pages={417--444},
  year={2016},
  publisher={Springer}
}

@inproceedings{le2025aircraft,
  title={Aircraft Resource-Constrained Assembly Line Balancing with Learning Effect: A Constraint Programming Approach},
  author={Le, Duc Anh and Roussel, St{\'e}phanie and Lecoutre, Christophe},
  booktitle={{31st International Conference on Principles and Practice of Constraint Programming (CP 2025)}},
  pages={25--1},
  year={2025},
  organization={Schloss Dagstuhl--Leibniz-Zentrum f{\"u}r Informatik}
}

@article{limbourg2012automatic,
  title={Automatic aircraft cargo load planning},
  author={Limbourg, Sabine and Schyns, Micha{\"e}l and Laporte, Gilbert},
  journal={Journal of the Operational Research Society},
  volume={63},
  number={9},
  pages={1271--1283},
  year={2012},
  publisher={Taylor \& Francis}
}

@inproceedings{pucel2024constraint,
  title={Constraint Programming Model for Assembly Line Balancing and Scheduling with Walking Workers and Parallel Stations},
  author={Pucel, Xavier and Roussel, St{\'e}phanie},
  booktitle={{30th International Conference on Principles and Practice of Constraint Programming}},
  year={2024}
}

@article{hubner2021solving,
  title={Solving the nuclear dismantling project scheduling problem by combining mixed-integer and constraint programming techniques and metaheuristics},
  author={H{\"u}bner, Felix and Gerhards, Patrick and St{\"u}rck, Christian and Volk, Rebekka},
  journal={Journal of Scheduling},
  volume={24},
  number={3},
  pages={269--290},
  year={2021},
  publisher={Springer}
}

@inproceedings{laborie2008reasoning,
  title={Reasoning with Conditional Time-Intervals.},
  author={Laborie, Philippe and Rogerie, Jerome},
  booktitle={{FLAIRS conference}},
  pages={555--560},
  year={2008}
}

@article{laborie2012interval,
  title={Interval-based language for modeling scheduling problems: An extension to constraint programming},
  author={Laborie, Philippe and Rogerie, J{\'e}r{\^o}me and Shaw, Paul and Vil{\'\i}m, Petr and Katai, Ferenc},
  journal={Algebraic Modeling Systems: Modeling and Solving Real World Optimization Problems},
  pages={111--143},
  year={2012},
  publisher={Springer}
}

@article{laborie2018ibm,
  title={{IBM} {ILOG} {CP} optimizer for scheduling: 20+ years of scheduling with constraints at {IBM/ILOG}},
  author={Laborie, Philippe and Rogerie, J{\'e}r{\^o}me and Shaw, Paul and Vil{\'\i}m, Petr},
  journal={Constraints},
  volume={23},
  pages={210--250},
  year={2018},
  publisher={Springer}
}

@article{Ozdamar1995,
    author = {Linet Özdamar and Gündüz Ulusoy},
    title = {A survey on the resource-constrained project scheduling problem},
    journal = {{IIE} Transactions},
    volume = {27},
    number = {5},
    pages = {574-586},
    year = {1995},
    publisher = {Taylor & Francis},
    doi = {10.1080/07408179508936773},
}

@article{Brucker1999,
	title = {Resource-constrained project scheduling: {Notation}, classification, models, and methods},
	volume = {112},
	issn = {0377-2217},
	doi = {10.1016/S0377-2217(98)00204-5},
	number = {1},
	journal = {European Journal of Operational Research},
	author = {Brucker, Peter and Drexl, Andreas and Möhring, Rolf and Neumann, Klaus and Pesch, Erwin},
	year = {1999},
	pages = {3--41},
}

@article{Hartmann2022,
	title = {An updated survey of variants and extensions of the resource-constrained project scheduling problem},
	volume = {297},
	issn = {0377-2217},
	doi = {10.1016/j.ejor.2021.05.004},
	number = {1},
	journal = {European Journal of Operational Research},
	author = {Hartmann, Sönke and Briskorn, Dirk},
	year = {2022},
	pages = {1--14},
}

@article{garey1975complexity,
  title={Complexity results for multiprocessor scheduling under resource constraints},
  author={Garey, Michael R and Johnson, David S.},
  journal={{SIAM} journal on Computing},
  volume={4},
  number={4},
  pages={397--411},
  year={1975},
  publisher={SIAM}
}

@inproceedings{bellenguez2004lower,
  title={Lower bounds for the multi-skill project scheduling problem with hierarchical levels of skills},
  author={Bellenguez, Odile and N{\'e}ron, Emmanuel},
  booktitle={{International Conference on the Practice and Theory of Automated Timetabling}},
  pages={229--243},
  year={2004},
  organization={Springer}
}

@inproceedings{young2017constraint,
  title={Constraint programming applied to the multi-skill project scheduling problem},
  author={Young, Kenneth D and Feydy, Thibaut and Schutt, Andreas},
  booktitle={{Principles and Practice of Constraint Programming: 23rd International Conference, CP 2017, Melbourne, Australia, August 28--September 1, 2017, Proceedings 23}},
  pages={308--317},
  year={2017},
  organization={Springer}
}

@article{shan2017adaptive,
  title={An adaptive genetic algorithm for demand-driven and resource-constrained project scheduling in aircraft assembly},
  author={Shan, Siqing and Hu, Zhongjun and Liu, Zhilian and Shi, Jihong and Wang, Li and Bi, Zhuming},
  journal={Information Technology and Management},
  volume={18},
  pages={41--53},
  year={2017},
  publisher={Springer}
}

@article{borsato2022integer,
title = {An integer programming mathematical model with line balancing and scheduling for standard work optimization: A realistic application to aircraft engines assembly lines},
journal = {Computers \& Industrial Engineering},
volume = {173},
pages = {108652},
year = {2022},
issn = {0360-8352},
doi = {10.1016/j.cie.2022.108652},
author = {Júlia da Matta Oliveira {Borsato Pinhão} and Anibal Alberto Vilcapoma Ignacio and Ormeu Coelho}
}

@article{niu2023short,
title = {Short-term aviation maintenance technician scheduling based on dynamic task disassembly mechanism},
journal = {Information Sciences},
volume = {629},
pages = {816-835},
year = {2023},
issn = {0020-0255},
doi = {10.1016/j.ins.2023.01.137},
author = {Ben Niu and Bowen Xue and Huifen Zhong and Haiyun Qiu and Tianwei Zhou}
}

@article{dayi2016lean,
  title={A Lean based process planning for aircraft disassembly},
  author={Dayi, O and Afsharzadeh, A and Mascle, Christian},
  journal={{IFAC-PapersOnLine}},
  volume={49},
  number={2},
  pages={54--59},
  year={2016},
  publisher={Elsevier}
}

@inproceedings{camelot2013decision,
  title={Decision support tool for the disassembly of reusable parts on an end-of-life aircraft},
  author={Camelot, Aurore and Baptiste, Pierre and Mascle, Christian},
  booktitle={{Proceedings of 2013 International Conference on Industrial Engineering and Systems Management (IESM)}},
  pages={1--8},
  year={2013},
  organization={{IEEE}}
}

@inproceedings{srinivasan1999selective,
  title={Selective disassembly of components with geometric constraints},
  author={Srinivasan, Hari and Gadh, Rajit},
  booktitle={{International Design Engineering Technical Conferences and Computers and Information in Engineering Conference}},
  volume={19746},
  pages={571--579},
  year={1999},
  organization={American Society of Mechanical Engineers}
}

@article{zhong2011disassembly,
  title={Disassembly sequence planning for maintenance based on metaheuristic method},
  author={Zhong, Lu and Youchao, Sun and Ekene Gabriel, Okafor and Haiqiao, Wu},
  journal={Aircraft Engineering and Aerospace Technology},
  volume={83},
  number={3},
  pages={138--145},
  year={2011},
  publisher={Emerald Group Publishing Limited}
}

@article{lee2002disassembly,
  title={Disassembly scheduling with capacity constraints},
  author={Lee, D-H and Xirouchakis, P and Zust, R},
  journal={{CIRP} Annals},
  volume={51},
  number={1},
  pages={387--390},
  year={2002},
  publisher={Elsevier}
}

@inproceedings{bentaha2013chance,
  title={Chance constrained programming model for stochastic profit--oriented disassembly line balancing in the presence of hazardous parts},
  author={Bentaha, Mohand Lounes and Batta{\"\i}a, Olga and Dolgui, Alexandre},
  booktitle={{Advances in Production Management Systems. Sustainable Production and Service Supply Chains: IFIP WG 5.7 International Conference, APMS 2013, State College, PA, USA, September 9-12, 2013, Proceedings, Part I}},
  pages={103--110},
  year={2013},
  organization={Springer}
}

@article{tian2013chance,
  title={A chance constrained programming approach to determine the optimal disassembly sequence},
  author={Tian, Guangdong and Zhou, MengChu and Chu, Jiangwei},
  journal={{IEEE} Transactions on Automation Science and Engineering},
  volume={10},
  number={4},
  pages={1004--1013},
  year={2013},
  publisher={{IEEE}}
}

@article{zwingmann2008optimal,
  title={Optimal disassembly sequencing strategy using constraint programming approach},
  author={Zwingmann, Xavier and Ait-Kadi, Daoud and Coulibaly, Amadou and Mutel, Bernard},
  journal={Journal of Quality in Maintenance Engineering},
  volume={14},
  number={1},
  pages={46--58},
  year={2008},
  publisher={Emerald Group Publishing Limited}
}

@article{edis2021constraint,
  title={Constraint programming approaches to disassembly line balancing problem with sequencing decisions},
  author={Edis, Emrah B},
  journal={Computers \& Operations Research},
  volume={126},
  pages={105111},
  year={2021},
  publisher={Elsevier}
}

@article{kizilay2022novel,
  title={A novel constraint programming and simulated annealing for disassembly line balancing problem with {AND/OR} precedence and sequence dependent setup times},
  author={Kizilay, Damla},
  journal={Computers \& Operations Research},
  volume={146},
  pages={105915},
  year={2022},
  publisher={Elsevier}
}

@article{kone2011event,
  title={Event-based {MILP} models for resource-constrained project scheduling problems},
  author={Kon{\'e}, Oumar and Artigues, Christian and Lopez, Pierre and Mongeau, Marcel},
  journal={Computers \& Operations Research},
  volume={38},
  number={1},
  pages={3--13},
  year={2011},
  publisher={Elsevier}
}

@article{Berthold2013,
	author = {Timo Berthold},
	doi = {https://doi.org/10.1016/j.orl.2013.08.007},
	issn = {0167-6377},
	journal = {Operations Research Letters},
	number = {6},
	pages = {611-614},
	title = {Measuring the impact of primal heuristics},
	volume = {41},
	year = {2013}
}

@inproceedings{thomas2024constraint,
  title={A Constraint Programming Approach for Aircraft Disassembly Scheduling},
  author={Thomas, Charles and Schaus, Pierre},
  booktitle={{International Conference on the Integration of Constraint Programming, Artificial Intelligence, and Operations Research}},
  pages={211--220},
  year={2024},
  organization={Springer}
}

@article{polo2023heuristic,
  title={Heuristic and metaheuristic methods for the multi-skill project scheduling problem with partial preemption},
  author={Polo-Mej{\'\i}a, Oliver and Artigues, Christian and Lopez, Pierre and M{\"o}nch, Lars and Basini, Virginie},
  journal={International Transactions in Operational Research},
  volume={30},
  number={2},
  pages={858--891},
  year={2023},
  publisher={Wiley Online Library}
}

@article{mischek2021local,
  title={A local search framework for industrial test laboratory scheduling},
  author={Mischek, Florian and Musliu, Nysret},
  journal={Annals of Operations Research},
  volume={302},
  number={2},
  pages={533--562},
  year={2021},
  publisher={Springer}
}

@article{laurent2017new,
  title={A new extension of the {RCPSP} in a multi-site context: Mathematical model and metaheuristics},
  author={Laurent, A and Deroussi, Laurent and Grangeon, Nathalie and Norre, Sylvie},
  journal={Computers \& Industrial Engineering},
  volume={112},
  pages={634--644},
  year={2017},
  publisher={Elsevier}
}

@article{sabaghi2016towards,
  title={Towards a sustainable disassembly/dismantling in aerospace industry},
  author={Sabaghi, Mahdi and Cai, Yongliang and Mascle, Christian and Baptiste, Pierre},
  journal={Procedia {CIRP}},
  volume={40},
  pages={156--161},
  year={2016},
  publisher={Elsevier}
}

@article{scheelhaase2022economic,
  title={Economic and environmental aspects of aircraft recycling},
  author={Scheelhaase, Janina and M{\"u}ller, Leon and Ennen, David and Grimme, Wolfgang},
  journal={Transportation Research Procedia},
  volume={65},
  pages={3--12},
  year={2022},
  publisher={Elsevier}
}

@article{asmatulu2013recycling,
  title={Recycling of Aircraft: State of the Art in 2011},
  author={Asmatulu, Eylem and Overcash, Michael and Twomey, Janet},
  journal={Journal of Industrial Engineering},
  volume={2013},
  number={1},
  pages={960581},
  year={2013},
  publisher={Wiley Online Library}
}

@article{habib2025current,
  title={Current Practices in Recycling and Reusing of Aircraft Materials and Equipment},
  author={Habib, Md Ahsan and Subeshan, Balakrishnan and Kalyanakumar, Chandrasekaran and Asmatulu, Ramazan and Rahman, Muhammad M and Asmatulu, Eylem},
  journal={Materials Circular Economy},
  volume={7},
  number={1},
  pages={1--36},
  year={2025},
  publisher={Springer}
}

@misc{kpmg2024circularity,
  title = {Circularity in flight},
  author = {{KPMG}},
  year = 2024,
  note = {\url{https://kpmg.com/ie/en/insights/aviation/circularity-in-flight-fs-aviation.html} [Accessed : 31 october 2025]}
}

@misc{gmi2024aircraft,
  title = {Aircraft Recycling Market - By Aircraft, Type, By Material \& Forecast, 2025 - 2034},
  author = {{Global Market Insights}},
  year = 2024,
  note = {\url{https://www.gminsights.com/industry-analysis/aircraft-recycling-market} [Accessed : 31 october 2025]}
}

@misc{fortune2025commercial,
  title = {Commercial Aircraft Disassembly, Dismantling and Recycling Market Size, Share \& COVID-19 Impact Analysis},
  author = {{Fortune Business Insights}},
  year = 2025,
  note = {\url{https://www.fortunebusinessinsights.com/commercial-aircraft-disassembly-dismantling-and-recycling-market-103584} [Accessed : 31 october 2025]}
}

@misc{iata2025global,
  title = {Global Outlook for Air Transport - June 2025},
  author = {{IATA}},
  year = 2025,
  note = {\url{https://www.iata.org/en/iata-repository/publications/economic-reports/global-outlook-for-air-transport-june-2025/} [Accessed : 31 october 2025]}
}

@misc{skybrary2025Aircraft,
  title = {Aircraft Ground Running},
  author = {{SKYbrary}},
  year = 2025,
  note = {\url{https://skybrary.aero/articles/aircraft-ground-running} [Accessed : 10 november 2025]}
}

@article{artigues2013note,
  title={A note on “event-based {MILP} models for resource-constrained project scheduling problems”},
  author={Artigues, Christian and Brucker, Peter and Knust, Sigrid and Kon{\'e}, Oumar and Lopez, Pierre and Mongeau, Marcel},
  journal={Computers \& Operations Research},
  volume={40},
  number={4},
  pages={1060--1063},
  year={2013},
  publisher={Elsevier}
}

@article{brimberg1996scheduling,
  title={Scheduling workers in a constricted area},
  author={Brimberg, J and Hurley, WJ and Wright, RE},
  journal={Naval Research Logistics ({NRL})},
  volume={43},
  number={1},
  pages={143--149},
  year={1996},
  publisher={Wiley Online Library}
}



\end{document}